\documentclass{article} 
\usepackage[final]{colm2026_conference}

\usepackage{microtype}
\usepackage{hyperref}
\usepackage{url}
\usepackage{booktabs}

\usepackage[T1]{fontenc}
\usepackage{lineno}

\usepackage{latexsym}
\usepackage{makecell}
\usepackage{amsmath}

\usepackage{inconsolata}
\usepackage{algorithm}
\usepackage{algorithmic}

\usepackage{graphicx}

\usepackage{hyperref}
\usepackage{url}
\usepackage{booktabs}

\usepackage{graphicx} 

\usepackage{multicol}
\usepackage{multirow}
\usepackage{graphicx}
\usepackage{xcolor}

\usepackage{listings}
\usepackage{lipsum} 

\newcommand\method{\textsc{Verb-ICL}}
\newcommand\votek{vote-\textit{k}}

\usepackage{amsmath,amssymb}

\usepackage{array}
\newcolumntype{C}[1]{>{\centering\arraybackslash}p{#1}}
\definecolor{customlightblue}{rgb}{0.25, 0.25, 1.0}
\definecolor{custompink}{rgb}{1.0, 0.1, 0.6} 

\definecolor{darkblue}{rgb}{0, 0, 0.5}
\hypersetup{colorlinks=true, citecolor=darkblue, linkcolor=darkblue, urlcolor=darkblue}

\title{\method{}: Rethinking In-Context Learning for Structured \mbox{Prediction}}

\author{Fan Bai$^{\clubsuit}$\thanks{Co-first author} \ \ Hengshuo Miao$^{\clubsuit*}$ \ Sanjit Singh Batra$^{\diamondsuit}$ \ Hamid Hassanzadeh$^{\diamondsuit}$ \\ \textbf{Ardavan Saeedi}$^{\diamondsuit}$ \ \textbf{Mark Dredze}$^{\clubsuit}$ \\
\textsuperscript{$\clubsuit$}\text{Johns Hopkins University} \ \textsuperscript{$\diamondsuit$}\text{Optum} \\
 \small \texttt{\{fbai3, hmiao5\}@jh.edu}, \ \texttt{mdredze@cs.jhu.edu}\ 
 \\
 }

\begin{document}

\ifcolmsubmission
\linenumbers
\fi

\maketitle

\begin{abstract}
Structured prediction tasks pose two challenges: outputs composed of sub-sentence units that require fine-grained local context understanding, and task-specific conventions that pretraining alone cannot supply. Both are especially difficult under few-shot in-context learning (ICL), where the model must infer them from a small demonstration set selectively annotated under a fixed budget. We propose \method{}, which addresses them along two axes: not only \emph{which} examples to annotate, but also \emph{what} those annotations should contain. \method{} selects examples with a token-level coverage strategy that captures the local patterns these tasks predict over, then generates error feedback that codifies these conventions and attaches it to the corresponding demonstrations. Across six datasets spanning information extraction and semantic parsing, and with multiple LLMs, \method{} consistently outperforms strong selective annotation baselines and continues to gain as the annotation budget increases. Analyses show that our feedback design beats globally aggregated and label-free alternatives, and that both components reduce structural violations. The feedback also helps regardless of the selection strategy.\footnote{Code is available at: \url{https://github.com/bflashcp3f/Verb-ICL}}
\end{abstract}

\section{Introduction}
\label{sec:intro}

Structured prediction tasks often involve complex, compositional outputs and require costly human annotation \citep{thompson1999active, hwa-2004, haffari-etal-2009-active}. The advent of large language models (LLMs; \citealp{brown2020languagemodelsfewshotlearners, touvron2023llamaopenefficientfoundation, dubey2024llama3herdmodels, openai2024gpt4ocard}) has introduced \textit{in-context learning} (ICL; \citealp{dong2024surveyincontextlearning}), a promising paradigm for these tasks. In ICL, all tasks, including structured prediction \citep{gupta-etal-2023-coverage}, are reformulated as generation problems, and the model performs inference by conditioning on a small set of labeled demonstrations without parameter updates. This setup is especially appealing when annotation is costly and task schemas evolve, as it can accommodate new schemas by updating demonstrations rather than retraining. However, the effectiveness of ICL depends critically on the choice of demonstration examples, making selective annotation under a fixed budget a central challenge.

\begin{figure*}[ht!]
    \centering
    \includegraphics[width=\textwidth]{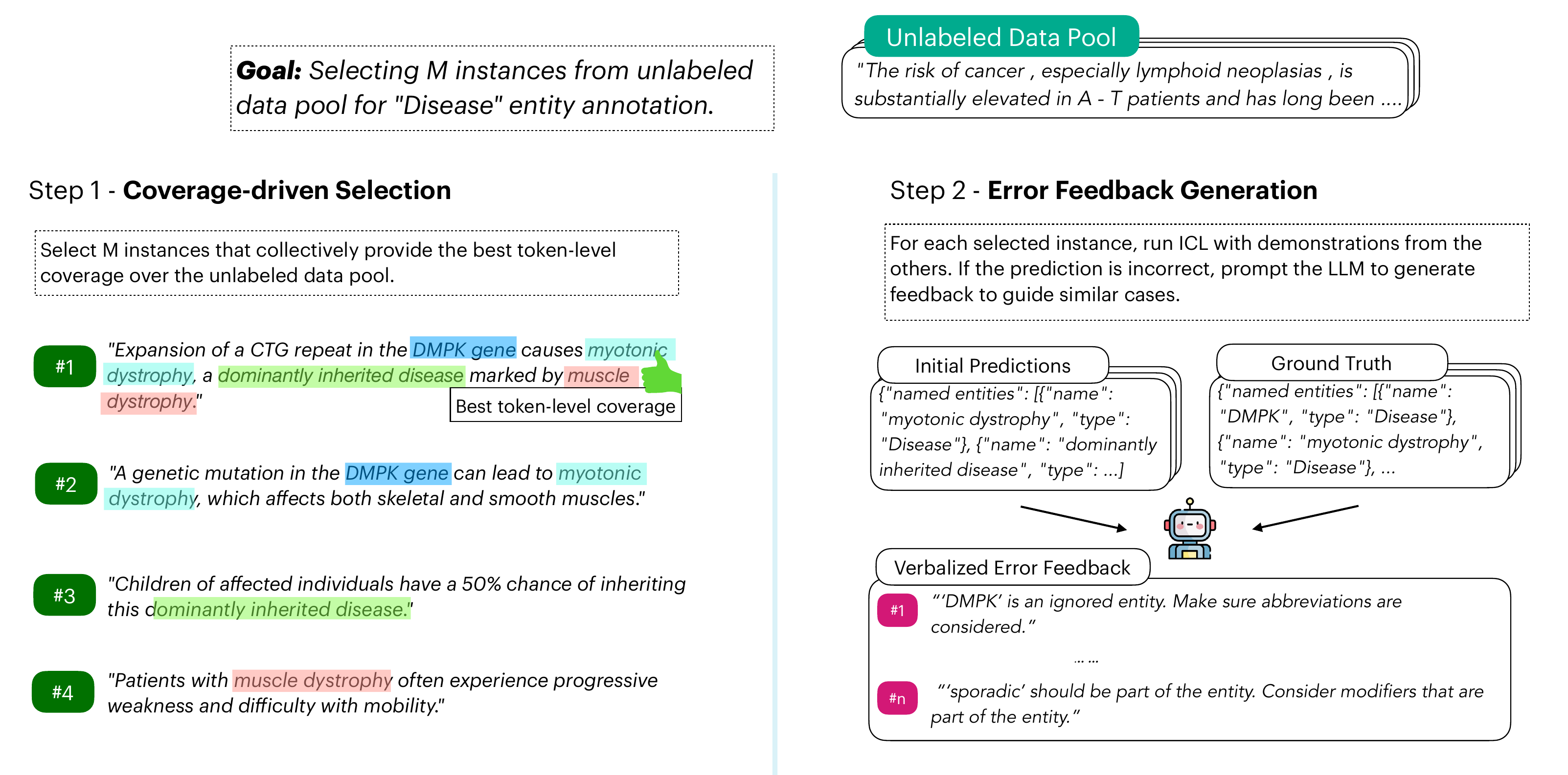}
    \vspace{-0.5cm}
    \caption{Overview of \method{}. In \textbf{Step 1}, we use a coverage-driven strategy to select $M$ examples that best cover the unlabeled pool $\mathcal{U}$. In \textbf{Step 2}, we collect model errors from annotated examples, and prompt the LLM to generate textual feedback that is incorporated as “Notes” in future demonstrations to guide subsequent predictions.
    }
    \label{fig:method}
    \vspace{-0.5cm}
\end{figure*}

Recent studies have explored selective annotation strategies tailored to in-context learning \citep{mavromatis2023examplesannotateincontextlearning, zhang2024ideal, mavromatis-etal-2024-covericl}. For example, \citet{su2023selective} and \citet{zhang2024ideal} construct directed graphs based on sentence embedding similarity and select examples that balance representativeness and informativeness. While such methods show promise for classification tasks, they are less suitable for structured prediction. First, sentence embeddings primarily capture global semantics and often miss the sub-sentence compositional patterns that drive prediction in tasks such as information extraction and semantic parsing, where different parts of the input may benefit from different demonstrations. More generally, selection should operate at the task's prediction unit, which is sub-sentence for the tasks we study, though it may be a full sentence or clause in tasks such as discourse parsing. Second, these approaches largely treat LLMs as statistical predictors, overlooking a key strength of LLMs: their ability to understand and generate natural language instructions. This is a missed opportunity for structured prediction, where task-specific conventions, such as entity boundary rules or logical form syntax, are human-defined artifacts that cannot be fully acquired through pretraining. Natural language feedback, akin to guidelines that human annotators provide on difficult cases, can help codify these conventions and complement standard input-output demonstrations.

To address these challenges, we propose \textbf{\method{}} (\textbf{Verb}alized \textbf{I}n-\textbf{C}ontext \textbf{L}earning), which rethinks how a fixed annotation budget is spent: not only \emph{which} examples to annotate, but also \emph{what} those annotations should contain. \method{} integrates fine-grained token-level coverage and reasoning-guided feedback into a unified pipeline. First, we select representative examples using a coverage-based strategy \citep{gupta-etal-2023-coverage}, ensuring broad coverage of local semantic patterns critical for structured prediction. Next, on this subset, we perform ICL inference to obtain initial predictions and explicitly prompt LLMs to generate natural language feedback on erroneous examples that explains the mistake and suggests actionable strategies for handling similar cases. These feedback notes are incorporated into the demonstration set, enriching ICL prompts with task-specific guidance.

We evaluate \method{} on six datasets spanning two structured prediction task types: information extraction and semantic parsing. Across annotation budgets from 16 to 512 examples, \method{} consistently outperforms seven selective annotation baselines under multiple LLMs. Feedback generated by models whose knowledge cutoff predates a dataset's release still improves performance on it, alleviating concerns about data contamination. Ablating the feedback design along four dimensions shows that these specific choices, rather than the idea of learning from errors alone, drive the gains. On the two semantic parsing datasets whose schemas define structural and value errors, structural errors dominate and both components reduce them. Finally, a manual audit finds 87--100\% of generated feedback useful, and feedback improves performance regardless of the selection strategy.

\section{Preliminaries: ICL-based Structured Prediction}
\label{sec:preliminaries}

ICL enables LLMs to perform unseen tasks by conditioning on a small set of labeled examples, without parameter updates \citep{brown2020languagemodelsfewshotlearners}. Formally, given a test input $x_{\text{test}}$ and a labeled set $\mathcal{L} = \{(x_1, y_1), \dots, (x_M, y_M)\}$ constructed in advance through selective annotation under a fixed budget $M$, a subset $\mathcal{D} \subseteq \mathcal{L}$ is selected as the demonstration set. This set is then used to construct a prompt for predicting the output $\hat{y}$ through the conditional probability $p(\hat{y} \mid x_{\text{test}}, \mathcal{D})$. Structured prediction poses particular challenges for ICL: the tasks we study predict over sub-sentence units, so different parts of an input may benefit from different demonstrations.

\paragraph{Coverage-Based Demonstration Selection}
The performance of ICL depends critically on the choice of demonstrations. A widely used approach is KATE \citep{liu-etal-2022-makes}, which selects in-context examples via nearest-neighbor search over sentence-level embeddings. However, sentence embeddings primarily capture global semantics and often miss fine-grained, local cues at the token level, such as entity boundary patterns or compositional sub-structures in logical forms. To address these issues, \citet{gupta-etal-2023-coverage} propose \textsc{SET-BSR}, a coverage-driven selection method that selects demonstrations to collectively maximize token-level coverage of the test instance.

Let $x = \langle t_1, \dots, t_{|x|} \rangle \in \mathcal{L}$ and $\hat{x} = \langle \hat{t}_1, \dots, \hat{t}_{|\hat{x}|} \rangle$ be a test instance, where each token is represented by a contextual embedding. \textsc{SET-BSR} proceeds in two steps:

\paragraph{Step 1: Token-level Coverage.}  
The coverage of a test token $\hat{t}_i$ by example $x$ is defined as:\footnote{IDF-weighted variants, $w(\hat{t}_i)\cdot \max_j \hat{t}_i^\top t_j$, have minimal impact on performance and are omitted, consistent with \citet{Zhang2020BERTScore}.}

\begin{equation}
\text{coverage}(\hat{t}_i, x) = \max_{1 \leq j \leq |x|} \hat{t}_i^\top t_j
\end{equation}

Aggregating these scores yields the sentence-level BERTScore-Recall (BSR; \citealp{Zhang2020BERTScore}):

\begin{equation}
\mathrm{BSR}(\hat{x}, x) = \sum_{\hat{t}_i \in \hat{x}} \text{coverage}(\hat{t}_i, x)
\end{equation}

For computational efficiency in the next step, token-level coverage scores for all labeled examples are organized into a \textbf{coverage matrix} $\mathbf{C} \in \mathbb{R}^{M \times d}$, where $C_{ij} = \text{coverage}(\hat{t}_j, x_i)$, $M = |\mathcal{L}|$ is the number of labeled examples, and $d = |\hat{x}|$ is the number of tokens in the test instance.

\paragraph{Step 2: Collective Coverage Maximization.}  The demonstration set $\mathcal{D}$ is constructed using a greedy algorithm (see Algorithm~\ref{alg:greedy}) that iteratively selects the instance with the largest marginal coverage gain from $\mathbf{C}$ until $k$ instances are chosen. Since the objective is submodular, this greedy strategy provides a constant-factor approximation guarantee \citep{nemhauser1978analysis}. As shown in Table~\ref{tab:demo_retrieval} in the appendix, SET-BSR consistently matches or outperforms sentence embedding–based retrievers. Therefore, we adopt it as the default retriever in ICL experiments.

\begin{figure}[t]
\centering
\setlength{\intextsep}{0pt}
\vspace{-0.7cm}
\begin{minipage}{0.75\textwidth}
\begin{algorithm}[H]
\caption{Greedy Optimization of Set Coverage}
\label{alg:greedy}
\small
\begin{algorithmic}[1]
\STATE \textbf{Input:} Budget $k$; coverage matrix $\mathbf{C} \in \mathbb{R}^{n \times d}$ ($n$ instances, $d$ tokens)
\STATE \textbf{Output:} Selected instance indices $\mathcal{Z}$
\STATE $\mathcal{Z} \gets \emptyset$
    \quad \textcolor{blue}{$\triangleright$ initialize selection}
\STATE $\mathbf{c} \gets -\infty$
    \quad \textcolor{blue}{$\triangleright$ initialize best coverage vector}
\WHILE{$|\mathcal{Z}| < k$}
    \STATE $\mathbf{C}' \gets \text{element\_max}(\mathbf{C},\, \mathbf{c})$
        \quad \textcolor{blue}{$\triangleright$ per-token best coverage in parallel}
    \STATE $\mathbf{s} \gets \text{sum}(\mathbf{C}',\, \text{axis}=1)$
        \quad \textcolor{blue}{$\triangleright$ total coverage score per instance}
    \STATE $\mathbf{s}[\mathcal{Z}] \gets -\infty$
        \quad \textcolor{blue}{$\triangleright$ exclude already selected}
    \STATE $z^* \gets \arg\max(\mathbf{s})$
        \quad \textcolor{blue}{$\triangleright$ highest-scoring instance}
    \STATE $\mathcal{Z} \gets \mathcal{Z} \cup \{z^*\}$
        \quad \textcolor{blue}{$\triangleright$ add to selection}
    \STATE $\mathbf{c} \gets \mathbf{C}'[z^*]$
        \quad \textcolor{blue}{$\triangleright$ update best coverage vector}
\ENDWHILE
\STATE \textbf{Return:} $\mathcal{Z}$
\end{algorithmic}
\end{algorithm}
\end{minipage}
\vspace{-0.4cm}
\end{figure}

\section{\method{}}
\label{sec:method}

We present \method{} (\textbf{Verb}alized \textbf{I}n-\textbf{C}ontext \textbf{L}earning), a method designed to maximize ICL performance on structured prediction tasks within a fixed annotation budget. As illustrated in Figure~\ref{fig:method}, \method{} consists of two key steps, corresponding to \emph{which} examples to annotate and \emph{what} those annotations contain: 1) coverage-driven selection to identify representative instances (\S\ref{sec:selection}), and 2) error feedback generation, which produces actionable guidance based on model errors (\S\ref{sec:ef}). We begin by outlining the problem setup, followed by detailed explanations of each step.

\paragraph{Problem Setup} 
We adopt the few-shot selective annotation setting used in prior work \citep{su2023selective, mavromatis-etal-2024-covericl, zhang2024ideal}. Given an unlabeled pool $\mathcal{U} = \{x_i\}_{i=1}^N$, the goal is to select a subset $\mathcal{L} \subset \mathcal{U}$ for annotation, subject to an annotation budget $|\mathcal{L}| = M$. The source datasets are fully labeled; we simulate annotation by hiding labels during selection and revealing the gold label of each selected example. These gold-labeled examples serve both as ICL demonstrations and as the reference for generating error feedback (\S\ref{sec:ef}). After annotation, for each test instance, a subset of labeled examples $\mathcal{D} \subset \mathcal{L}$ is retrieved as demonstrations using SET-BSR, as described in \S\ref{sec:preliminaries}.

\subsection{Coverage-driven Selection}
\label{sec:selection}

\method{} begins with a coverage-driven selection strategy inspired by \textsc{SET-BSR} (\S\ref{sec:preliminaries}; Step~1 in Figure~\ref{fig:method}). This design is motivated by the shared goal of both annotation selection and demonstration selection: identifying examples that collectively represent the target data distribution. The key distinction lies in their granularity, as demonstration selection targets a single test instance, while annotation selection operates over the entire unlabeled pool.

Building on this insight, we adapt \textsc{SET-BSR} for annotation selection by treating the unlabeled pool as a single aggregated “test instance” and selecting examples that maximize token-level coverage. For each unlabeled example $x_i$, we compute a coverage matrix $\mathbf{C}_i \in \mathbb{R}^{u \times d_i}$ over the unlabeled pool $\mathcal{U}$, where $u = |\mathcal{U}|$ is the number of unlabeled examples and $d_i = |x_i|$ is the number of tokens in $x_i$. These matrices are concatenated column-wise to form a global coverage matrix:

\begin{equation}
\small
\mathbf{C}_{\text{global}} = [\,\mathbf{C}_1 \,|\, \mathbf{C}_2 \,|\, \dots \,|\, \mathbf{C}_u\,] \in \mathbb{R}^{u \times (\sum_{i=1}^u d_i)}
\end{equation}

We apply the greedy algorithm in Algorithm~\ref{alg:greedy} to $\mathbf{C}_{\text{global}}$ to maximize coverage over the unlabeled pool. Since $\mathbf{C}_{\text{global}}$ can be large due to its token-level granularity, we adopt a simple and effective heuristic to reduce memory usage: for each $\mathbf{C}_i$, we retain only the top $K$ token columns (e.g., $K=10$), ranked by their aggregated coverage scores. As shown in Table~\ref{tab:coverage_max} in the appendix, this strategy reduces memory consumption with minimal impact on performance.

\subsection{Error Feedback Generation}
\label{sec:ef}

Unlike prior ICL-based selective annotation methods that stop at selecting labeled examples, \method{} takes a further step by prompting an LLM to generate natural language feedback on prediction errors, inspired by real-world annotation workflows where generalizable instructions on difficult cases complement ground-truth labels. This is especially valuable for structured prediction, where errors often stem from misunderstanding task-specific conventions, such as entity boundary rules or logical form syntax, that are human-defined artifacts not fully learnable from pretraining or input-output demonstrations alone.

As illustrated in Step~2 of Figure~\ref{fig:method}, we apply a leave-one-out strategy: each selected instance is treated as the query, and its demonstrations are retrieved from the remaining selected instances using SET-BSR. For those incorrectly predicted instances, we prompt the LLM to analyze the error and generate actionable feedback describing how similar cases should be handled. The models used for ICL inference and error feedback generation can differ; we analyze the effect of this choice in Section~\ref{sec:results}.

To ensure consistency and ease of integration, feedback is generated in a structured JSON format. During subsequent ICL inference, when feedback is available for a retrieved demonstration, the standard “Input–Output” format is expanded to an “Input–Notes–Output” format, where the “Notes” field contains the generated feedback. The full prompts and corresponding LLM responses are provided in Appendix~\ref{sec:prompt}. For demonstrations without feedback, the original format is retained. This error reflection process is conceptually related to recent work on prompt optimization \citep{pryzant-etal-2023-automatic, wang2023promptagent}. However, two key differences distinguish our approach. First, prompt optimization methods seek a single globally optimized prompt, whereas \method{} couples feedback with individual demonstrations, so different feedback is surfaced depending on which demonstrations are retrieved for a given query. Second, prompt optimization typically relies on a large held-out validation set to select the best prompt candidate; \method{} operates without such a set. In our implementation, we incorporate all generated feedback without filtering. As we show in Section~\ref{sec:results}, the strict per-example validation criterion is overly conservative: feedback that does not directly correct individual errors still captures generalizable task-level guidance and consistently improves overall performance.

\section{Experiment Setup}
\label{sec:experiment}

\subsection{Datasets and Evaluation Metric}

We conduct experiments on six structured prediction datasets spanning two task types. The first includes three information extraction datasets: NCBI \citep{dougan2014ncbi} and TweetNER7 \citep{ushio-etal-2022-named} for named entity recognition, and PHEE \citep{sun-etal-2024-leveraging} for event argument extraction. The second includes three semantic parsing datasets: GeoQuery \citep{geoquery}, ATIS \citep{hemphill-etal-1990-atis}, and MTOP \citep{li-etal-2021-mtop}. Dataset statistics (Table~\ref{tab:dataset_stats}) and detailed descriptions are provided in Appendix~\ref{sec:app_dataset}.\footnote{We use the English split of MTOP. Following \citet{gupta-etal-2023-coverage}, we reduce computational cost by randomly sampling 1{,}000 instances from the original test set.}

For evaluation, we use the standard mention-level matching metric \citep{tjong-kim-sang-de-meulder-2003-introduction} for the three information extraction datasets, where a prediction is considered correct only if both the entity or argument span boundary and its type exactly match the gold annotation. For the semantic parsing datasets, we follow \citet{gupta-etal-2023-coverage} and report exact match (EM) accuracy.

\subsection{Baselines}

We compare \method{} against seven baselines. \textbf{Random} and \textbf{K-Means} \citep{chang-etal-2021-training} serve as classic selection baselines. We further compare against three recent ICL-driven selective annotation methods: \textbf{\votek{}} \citep{su2023selective}, \textbf{IDEAL} \citep{zhang2024ideal}, and \textbf{CoverICL} \citep{mavromatis-etal-2024-covericl}. These methods construct similarity graphs over sentence embeddings and select examples that balance representativeness with informativeness, estimated via LLM confidence. We also include two active learning methods designed for generation tasks: \textbf{IDDS} \citep{tsvigun-etal-2022-active}, which balances diversity and dissimilarity from already selected instances, and \textbf{LDCAL} \citep{li-etal-2024-active}, which uses a certainty gain metric with outlier thresholding. Additional details are provided in Appendix~\ref{sec:app_baseline}.\footnote{Although fine-tuning is not the primary focus of this paper, we also explore applying \method{} to data selection for fine-tuning and observe promising results. Details are reported in Table~\ref{tab:finetune_comp}.}

\subsection{Implementation Details}
We conduct our main experiments using two LLMs for ICL inference: the proprietary \texttt{GPT-4o-mini} \citep{openai2024gpt4ocard} and the open-source \texttt{Qwen2.5-7B} \citep{qwen2025qwen25technicalreport}. For error feedback generation, we employ \texttt{GPT-4.1}, which provides more stable performance across tasks (see further discussion in \S\ref{sec:results}).\footnote{\texttt{gpt-4o-mini-2024-07-18} and \texttt{gpt-4.1-2025-04-14} are accessed via the OpenAI API.} We use identical prompts across inference LLMs and set the decoding temperature to 0 to achieve maximum reproducibility. Exact prompts are provided in Appendix~\ref{sec:prompt}. In all ICL experiments, each query retrieves eight demonstrations. Unless otherwise noted, every selection experiment is repeated with three random seeds, and we report the mean performance along with standard deviations. To establish strong and consistent baselines, we evaluate four embedding models commonly used in recent studies. As shown in Table~\ref{tab:emb_model}, \texttt{DeBERTa-Large-MNLI} achieves the best overall performance across datasets and is therefore adopted in all experiments.

\begin{table*}[!ht]
\centering
\small
\resizebox{0.75\linewidth}{!}{%
\begin{tabular}{clC{0.8cm}C{1.2cm}C{0.8cm}C{0.8cm}C{0.8cm}C{0.8cm}C{0.8cm}C{0.8cm}}
\toprule


\textbf{$|\mathcal{L}|$} & \textbf{Method} & NCBI & TwNER7 & PHEE & GeoQ & ATIS & MTOP  & Avg. \\

\midrule

\multirow{12}{*}{16} & Random & 61.2\textsubscript{4.4} & 52.3\textsubscript{1.4} & 56.6\textsubscript{2.7} & 23.0\textsubscript{3.8} & 27.3\textsubscript{0.8} & 11.9\textsubscript{2.1} & 38.7
\\ 

 & K-means  & 63.8\textsubscript{0.6} & 54.1\textsubscript{1.0} & 58.5\textsubscript{1.5} 
& 28.0\textsubscript{3.6}
 & 30.7\textsubscript{2.0} & 17.2\textsubscript{0.8} & 42.1
\\
 
 & IDDS \citeyearpar{tsvigun-etal-2022-active} & 65.0\textsubscript{0.3} & \textbf{55.4}\textsubscript{0.3} & 57.7\textsubscript{0.0} 
 & 27.7\textsubscript{0.2} & 35.6\textsubscript{0.3} & 15.3\textsubscript{0.2} & 42.8
\\
 
 & Vote-$k$ \citeyearpar{su2023selective} & 58.0\textsubscript{2.8} & 51.9\textsubscript{2.2} & 52.8\textsubscript{2.6} 
& 25.6\textsubscript{0.9} & 27.0\textsubscript{3.2} & 11.8\textsubscript{0.6} & 37.9
\\ 

  & \hspace{0.2cm} + SET-BSR & 59.4\textsubscript{3.0} & 52.9\textsubscript{1.6} & 54.6\textsubscript{1.9} 
& 29.3\textsubscript{2.6} & 27.7\textsubscript{4.1} & 14.7\textsubscript{2.0} & 39.8
\\

& LDCAL \citeyearpar{li-etal-2024-active} & 56.3\textsubscript{0.3} & 51.9\textsubscript{0.1} & 51.9\textsubscript{0.1}
& 16.3\textsubscript{0.2} & 16.0\textsubscript{0.0} & 9.1\textsubscript{0.1} & 33.6
\\

& CoverICL \citeyearpar{mavromatis-etal-2024-covericl} & 66.4\textsubscript{0.4} & 50.4\textsubscript{0.7} & 56.7\textsubscript{0.5} 
 & 29.2\textsubscript{0.4} & 29.8\textsubscript{1.3} & 13.3\textsubscript{0.8} & 41.0
\\
& \hspace{0.2cm} + SET-BSR & 
66.9\textsubscript{0.4} & 52.3\textsubscript{0.6} & 57.9\textsubscript{0.9} 
& 30.8\textsubscript{3.9} & 31.9\textsubscript{0.5} & 14.1\textsubscript{1.4} & 42.3
\\
 
 & IDEAL \citeyearpar{zhang2024ideal} & 62.7\textsubscript{3.3} & 50.9\textsubscript{1.0} & 54.0\textsubscript{1.4} 
& 23.8\textsubscript{1.0} & 26.7\textsubscript{1.2} & 9.7\textsubscript{2.1} & 38.0
\\

 & \hspace{0.2cm} + SET-BSR & 62.2\textsubscript{1.5} & 52.7\textsubscript{1.0} & 54.7\textsubscript{1.7} & 24.6\textsubscript{1.6} & 27.2\textsubscript{0.9} & 10.5\textsubscript{1.7} & 38.7
\\

 
 & \method{}\textsubscript{w/o EF} & 66.1\textsubscript{0.2}
 & 53.3\textsubscript{0.2} & 59.1\textsubscript{0.1} & 35.5\textsubscript{0.4}
 & 34.4\textsubscript{0.2} & 21.4\textsubscript{0.2}  
 & 45.0 \\ 

  & \method{} & \textbf{68.2}\textsubscript{0.3} & 53.6\textsubscript{0.1} & \textbf{59.5}\textsubscript{0.3} 
 & \textbf{36.3}\textsubscript{1.3} & \textbf{39.3}\textsubscript{0.2} & \textbf{22.5}\textsubscript{0.4} & \textbf{46.6}
\\
 
\midrule

\multirow{12}{*}{64} & Random & 67.7\textsubscript{0.3} & 55.3\textsubscript{0.9} & 60.2\textsubscript{0.8} 
& 53.7\textsubscript{0.7} & 42.5\textsubscript{2.7} & 24.7\textsubscript{1.2} & 50.7 \\ 

 & K-means  & 68.7\textsubscript{0.4} & 55.8\textsubscript{0.2} & 60.8\textsubscript{0.5} 
& 51.0\textsubscript{1.5} & 44.3\textsubscript{2.0} & 29.8\textsubscript{0.6} & 51.7
\\

 & IDDS \citeyearpar{tsvigun-etal-2022-active} & 67.7\textsubscript{0.1} & 54.8\textsubscript{0.2} & 60.4\textsubscript{0.1} 
 & 50.7\textsubscript{0.5} & 44.2\textsubscript{0.4} & 28.3\textsubscript{0.0} & 51.0
\\
 
 & Vote-$k$ \citeyearpar{su2023selective} & 66.8\textsubscript{0.1} & 52.9\textsubscript{0.6} & 59.7\textsubscript{0.4} 
 & 47.3\textsubscript{2.9} & 37.3\textsubscript{1.2} & 24.6\textsubscript{1.1} & 48.1
\\

  & \hspace{0.2cm} + SET-BSR & 67.8\textsubscript{1.2} & 53.7\textsubscript{0.4} & 60.7\textsubscript{0.7} 
& 51.0\textsubscript{3.2} & 42.6\textsubscript{0.5} & 27.9\textsubscript{0.9} & 50.6
\\ 

 & LDCAL \citeyearpar{li-etal-2024-active} & 61.0\textsubscript{0.3} & 53.0\textsubscript{0.1} & 58.9\textsubscript{0.1} 
 & 40.7\textsubscript{0.3} & 26.4\textsubscript{0.1} & 16.8\textsubscript{0.2} & 42.8
\\

& CoverICL \citeyearpar{mavromatis-etal-2024-covericl} & 66.9\textsubscript{0.9} & 54.1\textsubscript{0.2} & 58.6\textsubscript{0.5} 
 & 47.4\textsubscript{1.8} & 41.9\textsubscript{1.1}
& 23.9\textsubscript{1.4} & 48.8
\\
& \hspace{0.2cm} + SET-BSR & 
70.5\textsubscript{0.1} & 55.1\textsubscript{0.6} & 61.3\textsubscript{0.9} 
& 37.3\textsubscript{2.4} & 45.3\textsubscript{2.4} & 26.7\textsubscript{1.3} & 49.4
\\
 
 & IDEAL \citeyearpar{zhang2024ideal} & 64.4\textsubscript{0.7} & 53.7\textsubscript{1.2} & 58.8\textsubscript{0.7} 
& 48.1\textsubscript{1.1} & 38.4\textsubscript{1.7} & 22.5\textsubscript{0.3} & 47.7
\\

  & \hspace{0.2cm} + SET-BSR & 66.8\textsubscript{1.1} & 54.9\textsubscript{0.3} & 60.1\textsubscript{0.9} 
 & 50.6\textsubscript{3.4} & 43.5\textsubscript{1.4} & 24.4\textsubscript{0.4} & 50.0
\\ 

  
 & \method{}\textsubscript{w/o EF} & 69.4\textsubscript{0.0} & 55.6\textsubscript{0.1} & 61.1\textsubscript{0.1} &  53.1\textsubscript{0.3}
 & 48.7\textsubscript{0.5} & 33.8\textsubscript{0.2} 
 & 53.6 \\ 

 & \method{} & \textbf{71.3}\textsubscript{0.6} & \textbf{56.0}\textsubscript{0.5} & \textbf{62.1}\textsubscript{0.2} & 
\textbf{56.5}\textsubscript{0.4} & \textbf{53.0}\textsubscript{1.1} & \textbf{36.5}\textsubscript{0.9} & \textbf{56.0}
\\
 
\bottomrule
\end{tabular}%
}
\caption{Comparison of \method{} and seven baselines across six datasets using \texttt{GPT-4o-mini}, evaluated under annotation budgets of 16 and 64 examples. Results are averaged over three random seeds, with standard deviations reported as subscripts. For \votek{}, IDEAL and CoverICL, we also report results using their original sentence embedding similarity-based retrievers. \method{} outperforms baselines in both budget settings.}
\label{tab:main_results}
\vspace{-0.3cm}
\end{table*}

\section{Results and Analyses}
\label{sec:results}

\paragraph{Main Results} 
Table~\ref{tab:main_results} reports ICL performance on six datasets using \texttt{GPT-4o-mini} under annotation budgets of 16 and 64 examples. We compare \method{} against seven selective annotation baselines. \votek{}, IDEAL, and CoverICL are reported with their original sentence embedding–based retrievers and additionally paired with \textsc{SET-BSR} (\textbf{+ SET-BSR} rows); all other methods use \textsc{SET-BSR}. Even without error feedback, the coverage-driven selection component of \method{} outperforms baselines across both annotation budgets. Incorporating the error feedback step further improves \method{} performance, demonstrating the effectiveness of combining the two components. \votek{}, IDEAL, and CoverICL also benefit from being paired with \textsc{SET-BSR}, highlighting the importance of fine-grained, token-level coverage in demonstration retrieval. Table~\ref{tab:qwen_results} in the appendix presents results with \texttt{Qwen2.5-7B} on the same datasets and budgets, where the same trend holds.

To test whether token-level granularity alone suffices, and whether a recall-oriented objective simply favors longer examples, we compare two simpler variants in Appendix~\ref{sec:app_token_baselines}: BoW-coverage, which runs the same greedy coverage objective over bag-of-words vectors instead of contextual embeddings, and Longest-$n$, which selects the $M$ longest examples by word count. Both fall below \method{} on NCBI and ATIS, and Longest-$n$ falls below even Random, so contextual token representations rather than token-level granularity alone drive the gains. Length is likewise not the mechanism: \method{} selects far shorter examples than Longest-$n$ while outperforming it by 4.1 points, indicating that coverage rewards informativeness rather than length.

To verify that our findings generalize to more recent models, we repeat the comparison with \texttt{GPT-5.4-mini}\footnote{\texttt{gpt-5.4-mini-2026-03-17}.} on NCBI and ATIS, and find that the ordering above is unchanged (Table~\ref{tab:newer_llm}). Moreover, we show in Appendix~\ref{sec:app_ef_postprocessing} that error feedback also improves performance when applied on top of other selection strategies (Random, K-Means, and \votek{}), confirming that its benefits are orthogonal to the choice of selection method.

\begin{figure}[!t]
\begin{center}
  \includegraphics[width=\textwidth]
  {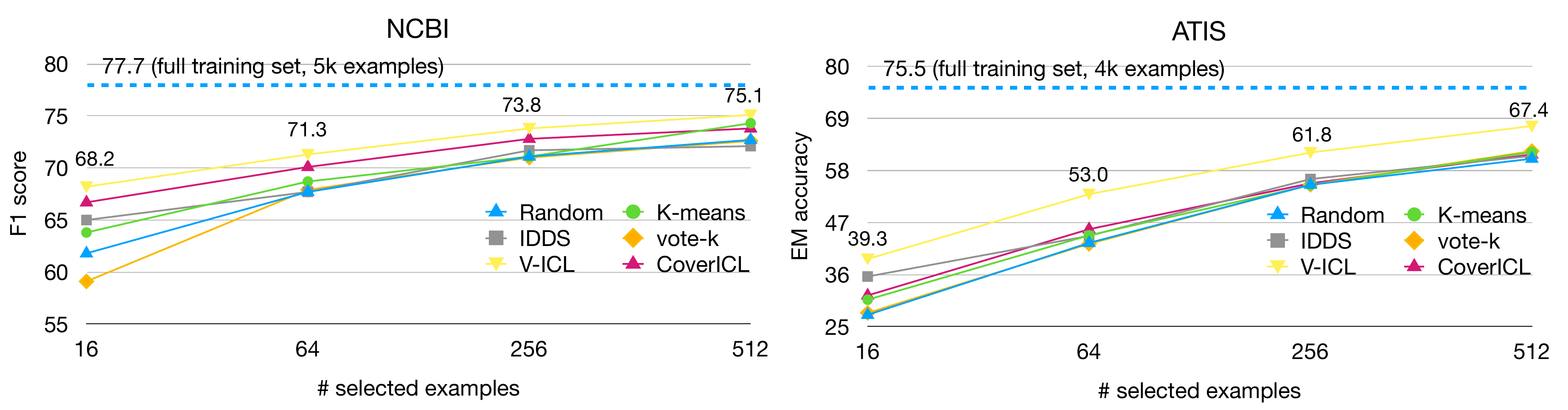}
  \vspace{-0.5cm}
\caption{Performance scaling with increasing annotation budgets from 16 to 512 examples on two representative datasets: NCBI (left) and ATIS (right). We compare \method{} with best-performing baselines: Random, K-Means, IDDS, \votek{}, and CoverICL. \method{} outperforms baselines at every budget level on both datasets.}
  \label{fig:scaling}
  \vspace{-0.3cm}
\end{center}
\end{figure}

\paragraph{Scaling Annotation Budget}
We scale the annotation budget from 16 to 512 examples to assess how different selection strategies perform under higher-resource conditions. We compare \method{} against the best baselines on NCBI and ATIS, as shown in Figure~\ref{fig:scaling}. Across all budget sizes, \method{} consistently achieves the best performance and approaches the full-data result on NCBI (over 5,000 examples) using only 512 labeled instances, highlighting the efficiency of selective annotation with \method{}.

\paragraph{Feedback Method Analysis}
The results above show that error feedback helps, but not how it should be incorporated, and prior work occupies different points in this space. We characterize the design along four dimensions. (1) \emph{Error source}: where the erroneous predictions come from, either zero-shot or leave-one-out ICL inference. (2) \emph{Grounding}: whether feedback is anchored in gold labels or in label-free self-critique. (3) \emph{Feedback selection}: which notes a query sees, all collected notes or only those attached to its retrieved demonstrations. (4) \emph{Positioning}: where the notes appear, in a global block at the top of the prompt or inline with each demonstration. Table~\ref{tab:feedback_design} compares six configurations, all using \method{}'s coverage-driven selection so that only the feedback design varies, and each adjacent pair isolating a single dimension. Row b1 instantiates the design of LEAP \citep{zhang2024leap}, and row e the label-free grounding of Self-Refine \citep{madaan2023selfrefine}. Appendix~\ref{sec:app_taxonomy} defines the dimensions in full and situates further methods within them.

\method{} (row d) performs best on both datasets, but the four dimensions contribute unevenly. Grounding matters most and most consistently: replacing gold-anchored feedback with label-free self-critique costs 1.7 points on NCBI and 5.0 on ATIS. Inline positioning gains 3.6 points on NCBI but is within noise on ATIS, and collecting errors from leave-one-out rather than zero-shot predictions helps NCBI (+1.9) while falling within noise on ATIS. Feedback selection is task-dependent: showing all notes is better on NCBI, while showing only the retrieved ones is better on ATIS. We adopt retrieval-selected feedback, which is also cheaper, as each query carries only the notes of its $K$ demonstrations rather than all $M$.

\begin{table}[h!]
\centering
\small
\begin{tabular}{lllllcc}
\toprule
& \multicolumn{4}{c}{\textbf{Feedback Design}} & \multicolumn{2}{c}{\textbf{Accuracy}} \\
\cmidrule(lr){2-5} \cmidrule(lr){6-7}
Method & Source & Grounding & Notes & Position & NCBI & ATIS \\
\midrule
(a) \method{} w/o EF & -- & -- & -- & -- & 69.6\textsubscript{0.1} & 48.7\textsubscript{0.5} \\
(b1) LEAP-style & zero-shot & gold & all & global & 68.0\textsubscript{0.9} & 50.4\textsubscript{0.8} \\
(b2) -- & LOO & gold & all & global & 69.9\textsubscript{0.2} & 49.5\textsubscript{1.5} \\
(c) -- & LOO & gold & retrieved & global & 67.7\textsubscript{0.4} & 52.1\textsubscript{0.5} \\
(d) \method{} & LOO & gold & retrieved & inline & \textbf{71.3}\textsubscript{0.7} & \textbf{52.9}\textsubscript{1.4} \\
(e) Self-Refine-style & LOO & label-free & retrieved & inline & 69.6\textsubscript{0.4} & 47.9\textsubscript{0.5} \\
\bottomrule
\end{tabular}
\caption{\label{tab:feedback_design} Error feedback design comparison on NCBI (mention-level F1) and ATIS (exact match), using \texttt{GPT-4o-mini} at a 64-example budget, averaged over three seeds. All rows use \method{}'s coverage-driven selection, so only the feedback design differs; EF denotes error feedback, and LOO means leave-one-out. Each adjacent pair isolates one dimension: error source (b1 vs.\ b2), feedback selection (b2 vs.\ c), positioning (c vs.\ d), and grounding (d vs.\ e).}
\end{table}

\paragraph{Feedback Quality and Filtering}
Given our design, a further question is whether all error-driven feedback should be kept. We approach this from two directions. First, Table~\ref{tab:feedback_analysis} in the appendix analyzes feedback produced by \texttt{GPT-4.1} across all six datasets under 16-shot and 64-shot budgets. We add a verification step that tests whether adding the feedback to the ICL prompt corrects the original error; we call such feedback \emph{verified}. Only a small fraction of feedback is verified, yet in 10 out of 12 settings, using all feedback outperforms using only verified feedback, indicating that feedback provides generalizable task-level guidance beyond instance-specific correction.

Second, we characterize the feedback directly. We pool items from all three seeds and manually label each item produced by \texttt{GPT-4.1} under the 64-shot setting into two useful categories (\emph{correct\_concrete}: specific, actionable rules tied to explicit constructs; \emph{correct\_general}: broader generalizable guidelines) and two harmful ones (\emph{misdiagnosis}: error type incorrectly identified; \emph{over-generalized/vague}: too broad to be actionable). Table~\ref{tab:feedback_quality} reports the distribution across all six datasets.

Overall, 87--100\% of the generated feedback falls into the two useful categories. Harmful feedback is rare and dataset-specific rather than systematic. Taken together, most feedback on these datasets is useful, and verification, the intuitive criterion, is too strict to improve on using all of it. We therefore apply no filtering, though better criteria may exist.

\begin{table}[h!]
\small
\centering
\resizebox{0.8\textwidth}{!}{
\begin{tabular}{lccccc}
\toprule
\textbf{Dataset} & \textbf{$n$} & \textbf{Concrete} & \textbf{General} & \textbf{Misdiag.} & \textbf{Over-gen./Vague} \\
\midrule
NCBI & 38 & 5 (13\%) & 28 (74\%) & 4 (11\%) & 1 (3\%) \\
TweetNER7 & 179 & 20 (11\%) & 151 (84\%) & 0 & 8 (4\%) \\
PHEE & 240 & 1 (0\%) & 225 (94\%) & 14 (6\%) & 0 \\
ATIS & 63 & 19 (30\%) & 43 (68\%) & -- & 1 (2\%) \\
GeoQuery & 66 & 48 (73\%) & 14 (21\%) & -- & 4 (6\%) \\
MTOP & 81 & 65 (80\%) & 16 (20\%) & -- & 0 \\
\bottomrule
\end{tabular}
}
\caption{\label{tab:feedback_quality}Distribution of feedback quality categories across all six datasets (64-shot). Misdiagnosis is verified for information extraction tasks only; ``--'' indicates not applicable for semantic parsing tasks. Over-generalized and vague categories are merged for compactness.}
\end{table}

\paragraph{What Errors Does \method{} Fix?}
To understand what \method{} improves, we examine its errors from the perspective of structured prediction, where they take two forms. A \emph{structural violation} breaks a task constraint, such as an invalid predicate, an arity mismatch, or a slot not licensed by the predicted intent; a \emph{value violation} leaves the structure intact and gets only a constant or span wrong \citep{pauk2026mapping, rajabyfaghihi2024consistent}. The distinction is defined by the output schema, so we run this analysis on ATIS and MTOP; entity spans in information extraction admit no comparable notion of structure. We classify each incorrect prediction with a rule-based parser that abstracts away constants and compares the remaining structure to gold; predictions that fail to parse, or are wrong in both respects, count as structural, so the two categories partition the errors. 

Table~\ref{tab:violation} confirms that structural violations dominate, at 81--86\% of errors under Random. Both components act on this same axis: coverage-driven selection reduces structural violations more than K-Means (MTOP $-17.7\%$ vs.\ $-8.8\%$), and error feedback deepens the reduction (ATIS $-13.6\%$ to $-18.5\%$; MTOP $-17.7\%$ to $-22.3\%$), consistent with feedback codifying conventions that are themselves structural. Selection and feedback therefore compound on the error type that dominates these tasks, rather than helping for unrelated reasons. Value counts on MTOP rise (104 to 151) as total errors fall (737 to 643), which is reclassification rather than new errors: a value violation is counted only when the structure is correct, so repairing structure moves predictions out of the structural category.

\begin{table}[h!]
\centering
\small
\begin{tabular}{lcccccc}
\toprule
& \multicolumn{3}{c}{\textbf{ATIS}} & \multicolumn{3}{c}{\textbf{MTOP}} \\
\cmidrule(lr){2-4} \cmidrule(lr){5-7}
\textbf{Method} & Total & Struct. & Value & Total & Struct. & Value \\
\midrule
Random & 608 & 491 & 117 & 737 & 633 & 104 \\
K-Means & 547 & 427 \small{(--13.0\%)} & 120 & 700 & 577 \small{(--8.8\%)} & 123 \\
\method{}\textsubscript{w/o EF} & 520 & 424 \small{(--13.6\%)} & 96 & 661 & 521 \small{(--17.7\%)} & 140 \\
\method{} & \textbf{487} & \textbf{400} \small{(--18.5\%)} & 87 & \textbf{643} & \textbf{492} \small{(--22.3\%)} & 151 \\
\bottomrule
\end{tabular}
\caption{\label{tab:violation} Error composition on ATIS and MTOP (\texttt{GPT-4o-mini}, 64-example budget, seed 42). Incorrect predictions are split into structural and value violations; parentheses give the relative reduction in structural violations against Random. Both components reduce structural violations; value counts rise because a value violation is counted only when the structure is correct.}
\end{table}

\paragraph{LLM Ablation and Data Contamination} 
Error feedback has so far been generated by \texttt{GPT-4.1}. Table~\ref{tab:ablation_llm} (left) instead compares \texttt{GPT-4o-mini}, \texttt{GPT-4o}, and \texttt{GPT-4.1} in their ability to generate error feedback on NCBI, ATIS, and PHEE, with \texttt{GPT-4o-mini} used as the ICL inference model under the 64-example budget. \texttt{GPT-4o-mini} and \texttt{GPT-4o} have a knowledge cutoff of October~1,~2023, which predates the release of the PHEE dataset in 2024. This setting therefore allows us to assess potential data contamination. On NCBI, all three models yield comparable improvements over the initial coverage-driven selection stage. In contrast, on ATIS and PHEE, \texttt{GPT-4o-mini} achieves only modest gains compared with \texttt{GPT-4o} and \texttt{GPT-4.1}, indicating that more capable LLMs exhibit stronger adaptability across diverse task types. Importantly, the improvements obtained by \texttt{GPT-4o-mini} and \texttt{GPT-4o} on PHEE, despite their earlier knowledge cutoff, alleviate concerns about data contamination.

\paragraph{LLM Cost}
Table~\ref{tab:ablation_llm} (right) reports the LLM cost of \method{} and two LLM-based baselines on NCBI under the 64-example setting.\footnote{\texttt{GPT-4o-mini}: \$0.15 per 1M input tokens and \$0.60 per 1M output tokens; \texttt{GPT-4.1}: \$2.00 per 1M input tokens and \$8.00 per 1M output tokens.} \method{} incurs LLM cost only for error feedback generation on selected examples, resulting in substantially lower total cost than \votek{} and CoverICL, which require inference over the entire selection pool.

\begin{table}[h!]
\small
\centering
\begin{minipage}[t]{0.52\textwidth}
\centering
\scalebox{0.82}{
\begin{tabular}{lcccc}
\toprule
 & & \multicolumn{3}{c}{\textbf{Error Feedback}} \\
 \cmidrule(l){3-5}
 &  & \texttt{4o-mini} & \texttt{4o} & \texttt{4.1} \\
\textbf{Dataset} & Cov. Sel. & (10/01/23) & (10/01/23) & (06/01/24) \\
\midrule
NCBI & 69.4\textsubscript{0.0} & 71.1\textsubscript{0.3} & 71.3\textsubscript{0.6} & 71.3\textsubscript{0.6} \\
ATIS & 48.7\textsubscript{0.5} & 49.7\textsubscript{0.4} & 51.2\textsubscript{0.1} & 53.0\textsubscript{1.1} \\
PHEE & 61.1\textsubscript{0.1} & 61.3\textsubscript{0.3} & 62.0\textsubscript{0.3} & 62.1\textsubscript{0.1}\\
\bottomrule
\end{tabular}
}
\end{minipage}%
\hfill
\begin{minipage}[t]{0.45\textwidth}
\centering
\scalebox{0.82}{
\begin{tabular}{lccc}
\toprule
\textbf{Method} & LLM & Selection & Feedback \\
\midrule
\votek{} & \texttt{4o-mini} & \$0.611 & -  \\
CoverICL & \texttt{4o-mini} & \$0.395 & -  \\
\midrule
\multirow{2}{*}{\method{}}  & \texttt{4o-mini} & - & \$0.033  \\
 & \texttt{4.1} &  & \$0.162 \\
\bottomrule
\end{tabular}
}
\end{minipage}
\caption{\label{tab:ablation_llm}\label{tab:cost} \textbf{Left:} LLM ablation for error feedback generation under the 64-example setting using \texttt{GPT-4o-mini} for ICL inference. Stronger models yield larger gains on complex tasks; gains on PHEE (released 2024) from pre-2024-cutoff models alleviate contamination concerns. \textbf{Right:} LLM cost comparison on NCBI (64-example). \method{} costs less by generating feedback only for coverage-selected examples.}
\end{table}

\section{Related Work}
\label{sec:related_work}

\paragraph{In-Context Learning for Structured Prediction}
In-context learning (ICL; \citealt{brown2020languagemodelsfewshotlearners}) is a representative few-shot learning paradigm that has attracted significant attention due to its strong cross-task generalization and sample efficiency. Recent studies \citep{pecher2024automaticcombinationsampleselection, baek2025revisitingincontextlearninglong, mao-etal-2025-survey} have investigated various approaches to improving ICL performance, from enhanced pre-training strategies \citep{chen-etal-2022-improving, gu-etal-2023-pre, shi2024incontext} to demonstration selection \citep{rubin-etal-2022-learning, levy-etal-2023-diverse}. Structured prediction is a demanding setting for this paradigm, since demonstrations must convey both fine-grained output structure and task-specific conventions. Prompted LLMs have been studied across named entity recognition \citep{monajatipoor2024llmsbiomedicinestudyclinical}, relation and event extraction \citep{wadhwa-etal-2023-revisiting, sun-etal-2024-leveraging}, semantic parsing \citep{shin-etal-2021-constrained}, and even table extraction \citep{bai-etal-2024-schema}. On the demonstration side, examples are selected by token-level coverage \citep{gupta-etal-2023-coverage} or label-grounded token statistics \citep{bai-etal-2025-llms}. All of this work takes a labeled pool as given; we revisit the setting under a fixed annotation budget, asking not only \emph{which} examples to annotate but also \emph{what} those annotations should contain.

\paragraph{Selective Annotation for ICL}
The data-scarce nature of few-shot ICL makes annotation selection critically important. This problem is closely related to active learning \citep{cohn1994improving, pool_al, settles2009active, iyer-etal-2017-learning, peris-casacuberta-2018-active, li-etal-2024-active, luo-etal-2023-weighting, zhang-etal-2024-elad, azeemi-etal-2025-label}, which has a long history in structured prediction \citep{thompson1999active, hwa-2004, haffari-etal-2009-active}. A key distinction is that selective annotation for ICL does not involve explicit model training, but instead relies on static, pretrained LLMs. Many recent works \citep{mavromatis-etal-2024-covericl, baumann2025largelanguagemodelhacking, xia-etal-2025-selection, berger-etal-2025-context} have explored this setting. For example, \citet{su2023selective} extend demonstration selection to the upstream task of selective annotation, constructing a directed graph over sentence embeddings to balance representativeness and informativeness, and \citet{zhang2024ideal} refine this graph-based formulation with an influence-driven criterion. In contrast, we draw inspiration from recent advances in demonstration selection \citep{gupta-etal-2023-coverage} and show that a token-level coverage-based strategy can be effectively extended to annotation selection, which is complementary to our use of LLM-generated error feedback.

\paragraph{Error Feedback}
A broad line of work uses LLM-generated natural language feedback on errors to improve downstream performance. Prompt optimization methods \citep{zhou2023largelanguagemodelshumanlevel, pryzant-etal-2023-automatic, wang2023promptagent, yuksekgonul2024textgradautomaticdifferentiationtext} iteratively refine a single global prompt by evaluating candidates on a held-out validation set. Self-Refine \citep{madaan2023selfrefine} and Reflexion \citep{shinn2023reflexion} instead critique and revise outputs at test time, while LEAP \citep{zhang2024leap} mines errors from zero-shot predictions and distills them into principles prepended to every prompt. \method{} differs on three axes. First, feedback is generated once at annotation time from the labeled examples, so no additional inference is required per test query. Second, each note stays attached to the demonstration that produced it, so different feedback surfaces depending on what is retrieved, rather than being aggregated into a single prompt. Third, \method{} requires no held-out validation set; as \S\ref{sec:results} shows, the per-example validation criterion used in prior work is too strict to improve on using all feedback.

\section{Conclusion}
\label{sec:conclusion}

We presented \method{}, which makes better use of a fixed annotation budget for ICL-based structured prediction along two axes: which examples to annotate, and what those annotations contain. Coverage-driven selection captures the fine-grained patterns these tasks predict over, while error feedback codifies task-specific conventions that cannot be acquired through pretraining or input-output demonstrations alone. Across six datasets, \method{} improves consistently over strong baselines, its specific feedback design matters, and both components reduce structural violations. These results suggest that when annotation is limited, what an annotation contains deserves as much attention as which examples are annotated.




\bibliography{colm2026_conference, custom}
\bibliographystyle{colm2026_conference}

\appendix

\clearpage

\section{Datasets}
\label{sec:app_dataset}

\begin{table}[h!]
\small
\begin{center}
\scalebox{0.9}{
\begin{tabular}{llc}
\toprule
 & & \textbf{\# Instances} \\
\textbf{Task Type} & \textbf{Dataset} & (train / dev / test) \\
\midrule
\multirow{3}{*}{Info. Extraction} & NCBI \citeyearpar{dougan2014ncbi} & 5.4k / 0.9k / 0.9k \\
 & TweetNER7 \citeyearpar{ushio-etal-2022-named} & 7.1k / 0.8k / 0.6k\\
 & PHEE \citeyearpar{sun-etal-2024-leveraging} & 2.9k / 1.0k / 1.0k \\
\midrule
\multirow{3}{*}{Semantic Parsing} & GeoQuery \citeyearpar{geoquery} & 0.7k / - / 0.2k \\
& ATIS \citeyearpar{hemphill-etal-1990-atis} & 4.0k / - / 1.0k \\
& MTOP \citeyearpar{li-etal-2021-mtop} &  15.7k / 2.2k / 4.4k \\

\bottomrule
\end{tabular}
}
\end{center}
\caption{\label{tab:dataset_stats} Statistics of the six evaluated datasets.
}
\end{table}

\paragraph{NCBI \citep{dougan2014ncbi}} 
The NCBI Disease dataset contains 793 PubMed abstracts with over 7,000 sentences. The task is to identify and extract disease entities, returning their exact text spans.

\paragraph{PHEE \citep{sun-etal-2024-leveraging}} 
The PHEE dataset contains over 5,000 annotated events extracted from medical case reports and biomedical literature. The task requires identifying 16 types of event arguments for two event categories: \emph{adverse events}, where drug usage causes harmful effects in a human patient, and \emph{potential therapeutic events}, where drug usage leads to beneficial effects.

\paragraph{TweetNER7 \citep{ushio-etal-2022-named}} 
The TweetNER7 dataset contains annotations for seven named entity types across 11{,}382 tweets collected from Twitter between September 2019 and August 2021. The entity types include \emph{group}, \emph{creative work}, \emph{person}, \emph{event}, \emph{product}, \emph{location}, and \emph{corporation}.

\paragraph{GeoQuery \citep{geoquery}} 
The GeoQuery dataset consists of 879 natural-language questions about U.S. geography, such as “Which states have cities named Austin?”. The task requires mapping each question to a corresponding logical form. 

\paragraph{ATIS \citep{hemphill-etal-1990-atis}} 
The ATIS dataset focuses on airline travel queries, where the goal is to parse natural-language utterances into logical forms expressed in lambda calculus. An example query is “What is the earliest flight from Washington to Atlanta leaving on Wednesday, September fourth?”, which can be represented as \texttt{(argmin \$0 (and (flight \$0) (from \$0 washington : ci) (to \$0 atlanta : ci) (day \$0 wednesday : da) (day\_number \$0 4 : dn) (month \$0 september : mn)) (departure\_time \$0))}.

\paragraph{MTOP \citep{li-etal-2021-mtop}} 
The MTOP dataset parses natural-language utterances into structured semantic frames consisting of an \emph{intent}, which captures the user’s overall goal, and multiple \emph{slots}, which represent specific arguments needed to fulfill that intent. For example, the utterance “I want to send Anthony Ford a video message.” is parsed as \texttt{[IN:SEND\_MESSAGE [SL:RECIPIENT Anthony Ford] [SL:TYPE\_CONTENT video]]}.

\section{Baselines}
\label{sec:app_baseline}



\paragraph{K-Means Clustering}
K-Means sampling selects representative instances by partitioning the embedding space into $k$ clusters. Each instance is first embedded into a $d$-dimensional space, after which K-Means is applied to identify cluster centroids. For each cluster, the instance with the highest cosine similarity to its centroid is selected, yielding a diverse set of representatives that approximates the global data distribution.


\paragraph{IDDS \citep{tsvigun-etal-2022-active}}
IDDS incrementally constructs a labeled set by balancing diversity within the remaining unlabeled pool and dissimilarity from already selected instances. At each iteration, candidates are scored using a weighted difference between their average similarity to unlabeled examples and their average similarity to labeled ones. This greedy strategy promotes broad in-domain coverage while reducing redundancy.

\paragraph{LDCAL \citep{li-etal-2024-active}}
LDCAL selects instances based on their marginal coverage gain beyond the current labeled set. It explicitly accounts for redundancy through the Coverage Excluding Redundancy (CER) score and evaluates candidates using the Average CER Gain (ACERG). A validity constraint ensures that selected instances contribute sufficient new coverage, yielding a principled balance between informativeness and redundancy control.

\paragraph{\votek{} \citep{su2023selective}}
\votek{} is a two-stage selection method that combines graph-based voting with LLM-based uncertainty estimation. In the first stage, a directed $k$-NN graph is constructed in the embedding space, where each node votes for its nearest neighbors. A greedy voting process selects a diverse seed set by favoring nodes supported by many distinct neighbors while penalizing redundant contributions. In the second stage, the seed set is used as in-context examples to estimate uncertainty over remaining candidates, and the remaining budget is allocated via stratified sampling across uncertainty bins to balance coverage and informativeness.

\paragraph{IDEAL \citep{zhang2024ideal}}
IDEAL formulates data selection as an influence maximization problem over a similarity graph. Examples are connected to their nearest neighbors with edge weights proportional to normalized similarity. Starting from a high-degree seed node, the algorithm greedily selects instances that maximize expected influence under the Independent Cascade diffusion model, estimated via Monte Carlo simulation. This approach explicitly models information propagation to identify globally influential examples.





\paragraph{CoverICL \citep{mavromatis-etal-2024-covericl}}
CoverICL targets difficult instances for in-context learning by explicitly covering hard examples in an embedding graph. After estimating uncertainty, the least confident half of the unlabeled pool is identified as the hard set. Each candidate defines a coverage region consisting of hard instances reachable within two hops in a $k$-NN graph. The final selection is obtained via greedy set cover, prioritizing examples that jointly maximize coverage of hard instances.







\section{\method{} Prompts}
\label{sec:prompt}

We use the NCBI dataset \citep{dougan2014ncbi}, which targets the identification of "Disease" entities, as an example to illustrate the prompts used in our experiments.

\textbf{In-Context Learning - Input}

\begin{lstlisting}
Here is the JSON template for named entity recognition:
{"named entities": [{"name": "ent_name_1", "type": "ent_type_1"}, ..., {"name": "ent_name_n", "type": "ent_type_n"}]}

Please identify "Disease" entities (exact text spans), following the JSON template above, and output the JSON object. If no named entities identified, output {"named entities": []}.

Input: In contrast , anti - Br ( a ) ( HPA - 5b ) only recognized the Lys ( 505 ) Thr ( 799 ) form , whereas anti - Br ( b ) ( HPA - 5a ) reacted with both Glu ( 505 ) Thr ( 799 ) and Glu ( 505 ) Met ( 799 ) isoforms .
Output: {"named entities": []}

Input: The closest proximal markers are APOC2 ( apolipoprotein C - II ) and CKM ( creatine kinase , muscle ) approximately 3 cM and 2 cM from the DM gene respectively , in the order APOC2 - CKM - DM .
Output: {"named entities": [{"name": "DM", "type": "Disease"}]}

Input: A 5 kb DNA sequence deletion was found to have its endpoints in the first and third introns .
Output: {"named entities": []}

Input: We have typed 22 breast and breast - ovarian cancer families with eight polymorphisms from the chromosome 17q12 - 21 region , including two in the EDH17B2 gene .
Output: {"named entities": [{"name": "breast and breast - ovarian cancer", "type": "Disease"}]}

Input: Chronic hemolysis was present in the absence of infections , oxidant drugs or ingestion of faba beans .
Output: {"named entities": [{"name": "Chronic hemolysis", "type": "Disease"}]}

Input: The present study supports the view that circulating lymphocytes give easy access to PAH gene transcripts whose nucleotide sequence is identical to that reported in liver and therefore represent a useful tool for molecular genetic studies in phenylketonuria . .
Output: {"named entities": [{"name": "phenylketonuria", "type": "Disease"}]}

Input: Fragments of the arylsulfatase A ( ARSA ) gene from a patient with juvenile - onset metachromatic leukodystrophy ( MLD ) were amplified by PCR and ligated into MP13 cloning vectors .
Output: {"named entities": [{"name": "metachromatic leukodystrophy", "type": "Disease"}, {"name": "MLD", "type": "Disease"}]}

Input: Mutations in this gene are responsible for Pendred syndrome and autosomal recessive non - syndromic hearing loss at the DFNB4 locus on chromosome 7q31 .
Output: {"named entities": [{"name": "Pendred syndrome", "type": "Disease"}, {"name": "autosomal recessive non - syndromic hearing loss", "type": "Disease"}, {"name": "DFNB4", "type": "Disease"}]}

Input: Terminal complement component deficiency predisposes to meningococcal infection and is inherited in an autosomal co - dominant manner .
Output:  
\end{lstlisting}

\textbf{In-Context Learning - Output}

\begin{lstlisting}
{"named entities": [{"name": "Terminal complement component deficiency", "type": "Disease"}]}
\end{lstlisting}

\textbf{Error Feedback - Input}

\begin{lstlisting}
Please evaluate the model's output for the task of identifying **"Disease" entities** by analyzing errors at the span level. Return the results in the structured JSON format shown below.

---

### Task Description:

- The task is to extract **Disease** entities from a sentence.
- Entities are matched based on their spans (token indices) in the sentence.
- Analyze each error independently without categorizing into predefined types.

---

### Evaluation Criteria:

1. **Overall Evaluation**:
   - Set `"Output Eval"` to `"Correct"` if and only if ALL entities are correctly identified with exact span boundaries.
   - Otherwise, set it to `"Incorrect"` and provide detailed error analysis.

2. **Span Matching**:
   - Two entities match if they have the same start and end token indices.
   - Entities with overlapping spans but different boundaries are considered errors.

3. **Error Analysis Guidelines**:
   - Analyze each error as a unique case without forcing it into predefined categories.
   - Each error should be reported as a separate entry with:
     - `error_id`: Integer starting from 0
     - `error_type`: One of "boundary_mismatch", "type_error", "false_positive", "false_negative"
     - `error_description`: A concise 1-2 sentence description explaining what the error is and why it likely occurred
     - `error_feedback`: A 1-2 sentence concrete rule or pattern instruction-following models should follow to make the correct prediction (e.g., "Include anatomical modifiers like 'breast' when they precede cancer types" or "Exclude generic terms like 'disease' unless they are part of a specific disease name")

---

### Input Format:

**Input Sentence:**  
`The present study supports the view that circulating lymphocytes give easy access to PAH gene transcripts whose nucleotide sequence is identical to that reported in liver and therefore represent a useful tool for molecular genetic studies in phenylketonuria . .`

**Ground Truth:**  
`{"named entities": [{"name": "phenylketonuria", "type": "Disease"}]}`

**Model Output:**  
`{"named entities": [{"name": "phenylketonuria", "type": "Disease"}]}`

**Span-Level Analysis:**
**Correct Predictions (Exact Matches):**
  [0] "phenylketonuria" <-> "phenylketonuria"

**Errors:**
  [No errors found]

---

### Output Format (JSON):

```json
{
  "Output Eval": "Correct" | "Incorrect",
  "Error Analysis": [
    {
      "error_id": 0,
      "error_type": "boundary_mismatch" | "type_error" | "false_positive" | "false_negative",
      "error_description": "<1-2 sentences: what is the error and why it occurred>",
      "error_feedback": "<1-2 sentences: specific feedback to fix the error and ensure correct predictions in the same context>"
    }
  ] | "NA"
}
```
\end{lstlisting}

\textbf{Error Feedback - Output}

\begin{lstlisting}
```json
{
  "Output Eval": "Incorrect",
  "Error Analysis": [
    {
      "error_id": 0,
      "error_type": "false_negative",
      "error_description": "The model failed to identify the Disease entity 'breast and breast - ovarian cancer', which is a specific disease name mentioned in the sentence. This likely occurred because the model did not recognize the full span of the entity due to its complexity.",
      "error_feedback": "Ensure that the model captures multi-part disease names accurately, especially when they include conjunctions or modifiers."
    }
  ]
}
```
\end{lstlisting}

\textbf{In-Context Learning w/ Error Feedback Notes - Input}

\begin{lstlisting}
Here is the JSON template for named entity recognition:
{"named entities": [{"name": "ent_name_1", "type": "ent_type_1"}, ..., {"name": "ent_name_n", "type": "ent_type_n"}]}

Please identify "Disease" entities (exact text spans), following the JSON template above, and output the JSON object. If no named entities identified, output {"named entities": []}.

Input: He was treated for club foot and hip dysplasia at birth .
Output: {"named entities": [{"name": "club foot", "type": "Disease"}, {"name": "hip dysplasia", "type": "Disease"}]}

Input: We postulate that altered metabolism of a natural toxin or antigen by the CYP2D6 gene may increase susceptibility to AS .
Output: {"named entities": [{"name": "AS", "type": "Disease"}]}

Input: Chronic hemolysis was present in the absence of infections , oxidant drugs or ingestion of faba beans .
Output: {"named entities": [{"name": "Chronic hemolysis", "type": "Disease"}]}

Input: We have typed 22 breast and breast - ovarian cancer families with eight polymorphisms from the chromosome 17q12 - 21 region , including two in the EDH17B2 gene .
Notes:
- Ensure that multi-part disease names are recognized as single entities, especially when they include conjunctions or modifiers.
Output: {"named entities": [{"name": "breast and breast - ovarian cancer", "type": "Disease"}]}

Input: Most patients have a heterozygous mutation in the APT1 gene , which encodes Fas ( CD95 , APO - 1 ) , mediator of an apoptotic pathway crucial to lymphocyte homeostasis .
Output: {"named entities": []}

Input: The present study supports the view that circulating lymphocytes give easy access to PAH gene transcripts whose nucleotide sequence is identical to that reported in liver and therefore represent a useful tool for molecular genetic studies in phenylketonuria . .
Output: {"named entities": [{"name": "phenylketonuria", "type": "Disease"}]}

Input: Terminal complement component deficiency predisposes to meningococcal infection and is inherited in an autosomal co - dominant manner .
Notes:
- Ensure that all disease entities mentioned in the sentence are captured, even if they are not the primary focus of the text.
Output: {"named entities": [{"name": "Terminal complement component deficiency", "type": "Disease"}, {"name": "meningococcal infection", "type": "Disease"}]}

Input: Fragments of the arylsulfatase A ( ARSA ) gene from a patient with juvenile - onset metachromatic leukodystrophy ( MLD ) were amplified by PCR and ligated into MP13 cloning vectors .
Notes:
- Focus on identifying the core disease name without including additional modifiers unless they are integral to the disease's recognized nomenclature.
- Ensure that common abbreviations for diseases are included in the training data and recognized as valid entities, especially when they are explicitly mentioned in the text.
Output: {"named entities": [{"name": "metachromatic leukodystrophy", "type": "Disease"}, {"name": "MLD", "type": "Disease"}]}

Input: Clustering of missense mutations in the ataxia - telangiectasia gene in a sporadic T - cell leukaemia .
Output: 
\end{lstlisting}

\textbf{In-Context Learning w/ Error Feedback Notes - Output}

\begin{lstlisting}
{"named entities": [{"name": "ataxia - telangiectasia", "type": "Disease"}, {"name": "T - cell leukaemia", "type": "Disease"}]}
\end{lstlisting}

\begin{table*}[!ht]

\centering
\small
\resizebox{0.75\linewidth}{!}{%
\begin{tabular}{clC{0.8cm}C{1.2cm}C{0.8cm}C{0.8cm}C{0.8cm}C{0.8cm}C{0.8cm}C{0.8cm}C{0.8cm}C{0.8cm}}
\toprule

\textbf{$|\mathcal{L}|$} & \textbf{Method} & NCBI & TwNER7 & PHEE & GeoQ & ATIS & MTOP & Avg. \\
\midrule

\multirow{9}{*}{16} & Random & 50.9\textsubscript{6.2}
 & 42.7\textsubscript{0.3} & 56.8\textsubscript{2.5} 
& 19.3\textsubscript{1.1} & 28.6\textsubscript{1.2} & 9.5\textsubscript{2.1} & 34.6
\\
& K-means & 52.0\textsubscript{1.8} & \textbf{44.0}\textsubscript{2.7} & 57.9\textsubscript{0.7} 
& 26.5\textsubscript{0.7} & 31.3\textsubscript{2.5} & 14.4\textsubscript{0.7} & 37.7
 \\
& IDDS \citeyearpar{tsvigun-etal-2022-active} & 45.0\textsubscript{0.0} & 42.2\textsubscript{0.3} & 58.1\textsubscript{0.0} 
& 24.5\textsubscript{0.2} & 38.9\textsubscript{0.0} & 12.4\textsubscript{0.0} & 36.9
 \\
& Vote-$k$ \citeyearpar{su2023selective} & 29.0\textsubscript{0.0} & 35.6\textsubscript{0.0}
 & 56.2\textsubscript{0.0} & 19.3\textsubscript{0.0} & 29.9\textsubscript{0.0} & 11.3\textsubscript{0.0} & 30.2
 \\
& LDCAL \citeyearpar{li-etal-2024-active} & 34.2\textsubscript{0.2} & 40.3\textsubscript{0.0} & 55.8\textsubscript{0.0} 
& 15.0\textsubscript{0.0} & 17.1\textsubscript{0.0} & 9.2\textsubscript{0.2} & 28.6
\\
& CoverICL \citeyearpar{mavromatis-etal-2024-covericl} & 
48.8\textsubscript{3.7} & 38.3\textsubscript{2.1} & 58.0\textsubscript{0.5} & 28.2\textsubscript{0.5} & 27.1\textsubscript{2.7} & 10.8\textsubscript{1.0} & 35.2
\\
& IDEAL \citeyearpar{zhang2024ideal} & 35.1\textsubscript{3.7} & 37.7\textsubscript{2.4} & 54.8\textsubscript{1.1} & 16.8\textsubscript{1.3} & 25.2\textsubscript{1.3} & 8.5\textsubscript{1.5} & 29.7
 \\


& \method{}\textsubscript{w/o EF} & 59.0\textsubscript{0.0} & 42.7\textsubscript{0.0} & 58.6\textsubscript{0.0} & 26.8\textsubscript{0.0} & 37.2\textsubscript{0.0} & 16.0\textsubscript{0.0} & 38.7
\\
 
& \method{} & \textbf{59.6}\textsubscript{1.5} & 43.3\textsubscript{1.2} & \textbf{58.7}\textsubscript{0.6} & \textbf{28.6}\textsubscript{1.7} & \textbf{37.4}\textsubscript{1.9} & \textbf{17.0}\textsubscript{0.5} & \textbf{40.8}
 \\
 
\midrule

 \multirow{9}{*}{64} & Random & 55.7\textsubscript{0.7} & 45.5\textsubscript{0.2} & 59.8\textsubscript{0.8}
& 46.2\textsubscript{1.1} & 42.0\textsubscript{2.1} & 20.9\textsubscript{0.5} & 45.0
 \\
 
& K-means & 56.4\textsubscript{1.8} & 45.1\textsubscript{0.8} & 59.4\textsubscript{0.8}
& 46.7\textsubscript{1.5} & 43.8\textsubscript{0.6} & 26.5\textsubscript{0.5} & 46.3
\\
 
& IDDS \citeyearpar{tsvigun-etal-2022-active} & 51.0\textsubscript{0.0} & 44.3\textsubscript{0.1} & 59.6\textsubscript{0.2}
& 46.5\textsubscript{0.2} & 43.4\textsubscript{0.0} & 24.1\textsubscript{0.1} & 44.8
 \\
 
& Vote-$k$ \citeyearpar{su2023selective} & 50.8\textsubscript{0.0} & 37.4\textsubscript{0.0}
 & 59.7\textsubscript{0.0} & 35.7\textsubscript{0.0} & 37.8\textsubscript{0.0} & 20.7\textsubscript{0.0} & 40.3
 \\

& LDCAL \citeyearpar{li-etal-2024-active} & 42.5\textsubscript{0.0} & 41.7\textsubscript{0.0} &  58.3\textsubscript{0.0}
& 35.7\textsubscript{0.0} & 25.5\textsubscript{0.0} & 15.4\textsubscript{0.1} & 36.5
 \\

& CoverICL \citeyearpar{mavromatis-etal-2024-covericl} & 
50.5\textsubscript{3.5} & 43.0\textsubscript{0.8}
 & 58.5\textsubscript{0.3} & 36.8\textsubscript{1.1} & 37.7\textsubscript{0.6} & 20.6\textsubscript{0.9} & 41.2
\\

& IDEAL \citeyearpar{zhang2024ideal} & 43.1\textsubscript{0.5}
 & 42.4\textsubscript{2.3} & 58.0\textsubscript{0.9} & 40.1\textsubscript{4.1} & 35.2\textsubscript{0.5} & 19.1\textsubscript{0.9} & 39.7
 \\


& \method{}\textsubscript{w/o EF} & 65.0\textsubscript{0.0} & 45.7\textsubscript{0.0} & 60.1\textsubscript{0.0} &  47.9\textsubscript{0.0} & 50.1\textsubscript{0.0} & 27.9\textsubscript{0.0} & 48.7
\\
 
& \method{}  & \textbf{66.4}\textsubscript{0.0} & \textbf{46.0}\textsubscript{1.2} & \textbf{60.1}\textsubscript{0.0} & \textbf{48.0}\textsubscript{1.9} & \textbf{51.0}\textsubscript{0.1} & \textbf{28.8}\textsubscript{0.3} & \textbf{50.1}
 \\

\bottomrule
\end{tabular}%
}
\caption{Comparison of \method{} and seven baselines across six datasets using \texttt{Qwen2.5-7B}, evaluated under two annotation budgets. \method{} outperforms baselines, demonstrating its robustness across different LLMs.}
\label{tab:qwen_results}
\end{table*}

\begin{table*}[!ht]

\centering
\small
\resizebox{0.9\linewidth}{!}{%
\begin{tabular}{llC{1cm}C{1cm}C{1cm}C{1.6cm}C{1cm}C{1cm}C{1cm}C{1cm}}
\toprule
& & \multicolumn{3}{c}{\textbf{Information Extraction}} & \multicolumn{3}{c}{\textbf{Semantic Parsing}} \\

\cmidrule(l){3-5} \cmidrule(l){6-8}

\textbf{LLM} & \textbf{Method} & NCBI & TwNER7 & PHEE & GeoQ & ATIS & MTOP & Avg. \\
\midrule

\multirow{3}{*}{\texttt{Qwen2.5-7B}} & Sen. Emb. & 66.7 & 45.0 & 61.7 & 63.2 & 67.7 & 61.1 & 60.9
\\
 & BSR & 68.7 & 47.2 & 62.7 & 66.4 & 72.7 & 64.3 & 63.6
\\
 & SET-BSR & 69.2 & 48.1 & 63.6 & 68.2 & 75.8 & 64.7 & 65.0 \\
 


\midrule

\multirow{3}{*}{\texttt{gpt-4o-mini}} & Sen. Emb. & 71.5 & 57.9 & 63.0 & 70.7 & 68.2 & 67.8 & 66.5 \\
 & BSR & 73.9 & 60.6 & 64.2 & 73.2 & 74.4 & 72.0 & 69.8 \\
 & SET-BSR & 77.7 & 60.8 & 64.7 & 75.0 & 75.5 & 72.2 & 71.0\\
 
\bottomrule
\end{tabular}%
}
\caption{Comparison of demonstration retrieval methods using the full training set. SET-BSR consistently outperforms sentence embedding-based retrievers across all settings.}
\label{tab:demo_retrieval}

\end{table*}

\begin{table*}[!ht]

\centering
\small
\renewcommand{\arraystretch}{1.05}
\resizebox{0.9\linewidth}{!}{%
\begin{tabular}{llC{1.5cm}C{1cm}C{1.2cm}C{1.5cm}C{1cm}C{1.2cm}}
\toprule
& & \multicolumn{3}{c}{\texttt{gpt-4o-mini}} & \multicolumn{3}{c}{\texttt{Qwen2.5-7B}} \\

\cmidrule(l){3-5} \cmidrule(l){6-8}

\textbf{Task} & \textbf{Dataset} & \textbf{$|\mathcal{L}|=16$} & Full & $\Delta$ & \textbf{$|\mathcal{L}|=16$} & Full & $\Delta$ \\ 
\midrule

\multirow{3}{*}{Information Extraction} & NCBI \citeyearpar{dougan2014ncbi} & 61.2\textsubscript{4.4} & 77.7 & +16.5 & 50.9\textsubscript{6.2} & 69.2 & +18.3
\\
& TweetNER7 \citeyearpar{ushio-etal-2022-named} & 52.3\textsubscript{1.4} & 60.8 & +8.5 & 42.7\textsubscript{0.3} & 48.1 & +5.4
\\
& PHEE \citeyearpar{sun-etal-2024-leveraging} & 56.6\textsubscript{2.7} & 64.7 & +8.1 & 56.8\textsubscript{2.5} & 63.6 & +6.8 \\

\midrule

\multirow{3}{*}{Semantic Parsing} & GeoQ \citeyearpar{geoquery} & 23.0\textsubscript{3.8} & 75.0 & +52.0 & 19.3\textsubscript{1.1} & 68.2 & +48.9
\\
& ATIS \citeyearpar{hemphill-etal-1990-atis} & 27.3\textsubscript{0.8} & 75.5 & +48.2 & 28.6\textsubscript{1.2} & 75.8 & +47.2
\\
& MTOP \citeyearpar{li-etal-2021-mtop} & 11.9\textsubscript{2.1} & 72.2 & +60.3 & 9.5\textsubscript{2.1} & 64.7 & +55.2\\







\bottomrule
\end{tabular}%
}

\caption{Comparison between randomly selecting 16 demonstrations and using the full training set with SET-BSR on six structured prediction datasets. Scaling from 16 to the full training set yields substantial improvements, confirming that these tasks benefit from additional annotations and more careful selection strategies.}
\label{tab:data_selection}

\end{table*}

\begin{table}[h!]
\small
\begin{center}
\scalebox{1.0}{
\begin{tabular}{ccc}
\toprule
\textbf{Dataset} & \textbf{\# token/sent.} & Results \\
\midrule
\multirow{4}{*}{NCBI} & 5 & 68.9\textsubscript{0.5}  \\
 & 10 & 70.7\textsubscript{0.2} \\
 & 15 & 68.6\textsubscript{0.2} \\
 & full & 69.6\textsubscript{0.1} \\
\midrule
\multirow{4}{*}{GeoQuery} & 5 & 51.6\textsubscript{0.5} \\
 & 10 & 55.1\textsubscript{0.5} \\
 & 15 & 51.8\textsubscript{0.6} \\
 & full & 54.3\textsubscript{0.7} \\
\bottomrule
\end{tabular}
}
\end{center}
\caption{\label{tab:coverage_max} \method{} coverage-driven selection results (64-example budget) with different column sizes for $\mathbf{C}_i$. We can see that reducing the column size to 10 does not impact the performance of coverage-driven selection significantly.
}
\end{table}

\begin{table}[h!]
\small
\begin{center}
\scalebox{1.0}{
\begin{tabular}{lccc}
\toprule
\textbf{Method} & 16-shot & 64-shot & 256-shot \\
\midrule
 Label Reweight & 2.5 & 8.3 & 47.8 \\
 \method{} & 30.3 & 44.4 & 52.1 \\
\bottomrule
\end{tabular}
}
\end{center}
\caption{\label{tab:finetune_comp} Fine-tuning BERT-base performance with data selected by \method{} vs. Label Reweight \citep{luo-etal-2023-weighting} on CoNLL03. Across all three annotation budgets (16, 64, and 256 examples), data selected by \method{} consistently leads to higher fine-tuning performance.
}
\end{table}

\begin{table*}[!ht]

\centering
\small
\resizebox{0.9\linewidth}{!}{%
\begin{tabular}{clC{1.6cm}C{1.6cm}C{1.6cm}C{1.6cm}C{1.6cm}}
\toprule


\textbf{$|\mathcal{L}|$} & \textbf{Model} & NCBI  & PHEE & GeoQ & Average \\
\midrule

\multirow{4}{*}{16} & DeBERTa-Large-MNLI & 66.2\textsubscript{0.2} & 59.1\textsubscript{0.1} & 35.5\textsubscript{0.4} & 53.6
 \\
& DeBERTa-Base-MNLI & 65.4\textsubscript{0.2} & 57.4\textsubscript{0.2} & 34.9\textsubscript{0.6} & 52.6 \\
& BERT-base-uncased & 64.3\textsubscript{0.2} & 56.7\textsubscript{0.2} & 39.2\textsubscript{0.4} & 53.4 \\
& all-mpnet-base-v2 & 63.2\textsubscript{0.2} & 57.3\textsubscript{0.2} & 34.0\textsubscript{0.2} & 51.5
\\

\midrule

 \multirow{4}{*}{64} & DeBERTa-Large-MNLI & 69.7\textsubscript{0.1} & 61.1\textsubscript{0.1} & 53.1\textsubscript{0.3} & 61.3
 \\
 & DeBERTa-Base-MNLI & 67.8\textsubscript{0.2} & 61.1\textsubscript{0.2} & 54.3\textsubscript{0.3} & 61.1\\
& BERT-base-uncased & 71.5\textsubscript{0.2} & 59.6\textsubscript{0.3} & 54.9\textsubscript{0.3} & 62.0 \\
& all-mpnet-base-v2 & 68.8\textsubscript{0.1} & 58.4\textsubscript{0.1} & 53.5\textsubscript{0.4} & 60.2
\\
\bottomrule
\end{tabular}%
}

\caption{Comparison of four embedding models for ICL using \texttt{GPT-4o-mini}. \texttt{DeBERTa-Large-MNLI} demonstrates great performance in most settings.}
\label{tab:emb_model}
\end{table*}

\begin{table*}[h!]
\small
\centering
\scalebox{0.9}{
\begin{tabular}{lcccccccccccc}
\toprule
 & \multicolumn{2}{c}{\textbf{NCBI}} & \multicolumn{2}{c}{\textbf{ATIS}} & \multicolumn{2}{c}{\textbf{GeoQuery}} & \multicolumn{2}{c}{\textbf{TweetNER7}} & \multicolumn{2}{c}{\textbf{PHEE}} & \multicolumn{2}{c}{\textbf{MTOP}} \\
 \cmidrule(lr){2-3} \cmidrule(lr){4-5} \cmidrule(lr){6-7} \cmidrule(lr){8-9} \cmidrule(lr){10-11} \cmidrule(lr){12-13}
 & 16 & 64 & 16 & 64 & 16 & 64 & 16 & 64 & 16 & 64 & 16 & 64 \\
\midrule
\#Total feedback        & 8  & 26 & 15 & 44 & 12 & 43 & 16 & 60 & 14 & 60 & 14 & 53 \\
\#Verified feedback      & 4  & 14 & 2  & 17 & 4  & 15 & 7  & 15 & 0  & 4  & 2  & 16 \\
ICL w/ verified only   & 67.4 & 70.6 & 36.0 & 50.8 & 37.5 & 54.3 & 55.4 & 55.0 & 58.9 & 61.4 & 21.7 & 35.2 \\
ICL w/ all feedback    & 68.6 & 71.1 & 39.1 & 51.7 & 35.4 & 56.1 & 53.6 & 55.6 & 59.6 & 61.8 & 22.7 & 35.7 \\
\bottomrule
\end{tabular}
}
\caption{\label{tab:feedback_analysis}Analysis of LLM-generated error feedback across all six datasets under 16-shot and 64-shot annotation budgets, using a single seed (42) to reduce cost, as variance across seeds is small. ``Verified'' feedback refers to feedback that directly corrects the original error when added to the ICL prompt. Using all feedback beats using only verified feedback in 10 of 12 settings, so feedback provides task-level guidance beyond instance-specific correction.}
\end{table*}

\section{Error Feedback as Post-Processing}
\label{sec:app_ef_postprocessing}

\begin{table}[h!]
\small
\centering
\scalebox{0.9}{
\begin{tabular}{lcccc}
\toprule
 & \multicolumn{2}{c}{\textbf{NCBI}} & \multicolumn{2}{c}{\textbf{ATIS}} \\
 \cmidrule(lr){2-3} \cmidrule(lr){4-5}
\textbf{Method} & 16-shot & 64-shot & 16-shot & 64-shot \\
\midrule
Random & 67.1 & 68.6 & 28.5 & 39.7 \\
\hspace{0.2cm} + EF & 67.6\textsubscript{+0.5} & 69.3\textsubscript{+0.7} & 29.4\textsubscript{+0.9} & 42.9\textsubscript{+3.2} \\
\midrule
K-Means & 63.4 & 69.3 & 28.2 & 45.7 \\
\hspace{0.2cm} + EF & 65.7\textsubscript{+2.3} & 71.4\textsubscript{+2.1} & 28.4\textsubscript{+0.2} & 48.8\textsubscript{+3.1} \\
\midrule
\votek{} & 64.1 & 66.7 & 29.6 & 42.5 \\
\hspace{0.2cm} + EF & 65.4\textsubscript{+1.3} & 70.4\textsubscript{+3.7} & 30.3\textsubscript{+0.7} & 44.8\textsubscript{+2.3} \\
\bottomrule
\end{tabular}
}
\caption{\label{tab:ef_postprocessing} Effect of applying error feedback (EF) as a post-processing step on top of different selection strategies, computed with \texttt{GPT-4o-mini} at a single seed (42). The claim that feedback helps regardless of selection method rests on the within-row EF gain, which compares the same selected examples with and without feedback and is unaffected by the selection seed.}
\end{table}

To assess whether the benefits of error feedback are specific to our coverage-driven selection strategy, we apply error feedback (EF) as a post-processing module on top of three alternative selection methods: Random, K-Means, and \votek{}. Table~\ref{tab:ef_postprocessing} reports results on NCBI and ATIS under 16-shot and 64-shot annotation budgets. Error feedback consistently improves all 12 settings, with gains ranging from +0.2 to +3.7 points. The improvements are particularly pronounced in the 64-shot setting, where more annotated examples provide richer error patterns for feedback generation. These results demonstrate that error feedback is orthogonal to the choice of selection strategy: it optimizes \emph{how} examples are presented to the model, not \emph{which} examples are selected.

\section{Token-level Selection Baselines}
\label{sec:app_token_baselines}

\method{} selects examples using contextual token-level coverage. Two simpler strategies test whether this is necessary. \textbf{BoW-coverage} applies the same greedy coverage objective of Algorithm~\ref{alg:greedy}, but represents each example as a bag-of-words vector rather than a sequence of contextual embeddings, so coverage is measured over vocabulary overlap alone. \textbf{Longest-$n$} ignores coverage entirely and selects the $M$ examples with the most words. Both use \textsc{SET-BSR} for demonstration retrieval at test time, as in all other experiments, so only the annotation selection differs.

Table~\ref{tab:token_baselines} reports accuracy alongside the average length of the selected examples. Neither simpler strategy matches \method{}, and the margin widens from NCBI to ATIS, where compositional logical forms require distinguishing between structurally similar inputs that share vocabulary. Longest-$n$ falls below Random on both datasets despite selecting the longest examples available, which indicates that the recall-oriented coverage objective does not succeed by favoring length: \method{} selects examples roughly half as long as Longest-$n$ and outperforms it by 4.1 and 11.1 points.

\begin{table}[h!]
\centering
\small
\begin{tabular}{lcccc}
\toprule
& \multicolumn{2}{c}{\textbf{Accuracy}} & \multicolumn{2}{c}{\textbf{Avg.\ length}} \\
\cmidrule(lr){2-3} \cmidrule(lr){4-5}
\textbf{Method} & NCBI & ATIS & NCBI & ATIS \\
\midrule
Random & 67.7\textsubscript{0.3} & 42.5\textsubscript{2.7} & 23.8 & 10.9 \\
BoW-coverage & 68.6\textsubscript{0.2} & 41.0\textsubscript{0.1} & 64.7 & 19.3 \\
Longest-$n$ & 65.5\textsubscript{0.3} & 37.6\textsubscript{0.1} & 75.5 & 27.4 \\
\method{}\textsubscript{w/o EF} & \textbf{69.6}\textsubscript{0.1} & \textbf{48.7}\textsubscript{0.5} & 42.0 & 13.1 \\
\bottomrule
\end{tabular}
\caption{\label{tab:token_baselines} Token-level selection baselines at a 64-example budget with \texttt{GPT-4o-mini}, averaged over three seeds. NCBI reports mention-level F1 and ATIS exact match. Average length is the word count of the $M$ selected examples after whitespace tokenization and lowercasing.}
\end{table}



\begin{table}[h!]
\centering
\small
\begin{tabular}{lcccc}
\toprule
& \multicolumn{2}{c}{\textbf{NCBI}} & \multicolumn{2}{c}{\textbf{ATIS}} \\
\cmidrule(lr){2-3} \cmidrule(lr){4-5}
\textbf{Method} & 16 & 64 & 16 & 64 \\
\midrule
Random & 68.6\textsubscript{2.9} & 73.5\textsubscript{1.2} & 33.2\textsubscript{0.8} & 50.8\textsubscript{3.3} \\
K-Means & 70.7\textsubscript{0.7} & 74.5\textsubscript{1.0} & 36.3\textsubscript{1.9} & 52.0\textsubscript{2.7} \\
IDDS & 72.8\textsubscript{0.9} & 72.9\textsubscript{0.3} & 41.2\textsubscript{0.8} & 53.1\textsubscript{0.4} \\
\method{}\textsubscript{w/o EF} & 73.2\textsubscript{0.4} & 75.7\textsubscript{0.4} & 45.9\textsubscript{0.3} & 58.7\textsubscript{0.5} \\
\method{} & \textbf{74.7}\textsubscript{0.1} & \textbf{76.9}\textsubscript{0.3} & \textbf{46.0}\textsubscript{1.2} & \textbf{59.7}\textsubscript{0.7} \\
\bottomrule
\end{tabular}
\caption{\label{tab:newer_llm} ICL performance with \texttt{GPT-5.4-mini} under 16- and 64-example annotation budgets, averaged over three seeds. NCBI reports mention-level F1 and ATIS exact match. Feedback is generated by \texttt{GPT-4.1}, as in the main experiments.}
\end{table}

\section{A Design Space for Error Feedback}
\label{sec:app_taxonomy}

Methods that learn from model errors bundle together several independent choices, which makes them difficult to compare directly. We describe the four dimensions used in \S\ref{sec:results} and situate existing methods within them.

\paragraph{Error source.} Where the errors used to generate feedback come from. LEAP \citep{zhang2024leap} induces mistakes by prompting the model zero-shot on the few available examples. \method{} collects errors under leave-one-out few-shot inference, so each prediction is made under conditions closer to test time, where demonstrations are always present. Self-Refine \citep{madaan2023selfrefine} and Reflexion \citep{shinn2023reflexion} instead collect errors at test time, from the model's own output on the query being answered.

\paragraph{Grounding.} What the feedback is anchored to. Gold-grounded feedback compares a prediction against its reference label, so the model is shown what the correct output was and can diagnose the gap. Label-free feedback asks the model to critique its own output without a reference. Grounding is available only when errors are collected on labeled data, so test-time methods are necessarily label-free.

\paragraph{Feedback selection.} Which notes a given query sees. Global approaches show every collected note to every query. Retrieval-selected approaches show only the notes attached to the demonstrations retrieved for that query, so the notes vary across queries and the added prompt length is bounded by the retrieval depth $K$ rather than by the size $M$ of the annotated set.

\paragraph{Positioning.} Where the notes appear in the prompt. A global block places all notes together, typically as a prefix preceding the demonstrations. Inline positioning places each note with the demonstration it was derived from, preserving the association between an error and the example that produced it.

Table~\ref{tab:design_space} situates existing methods across these dimensions. The last two dimensions presuppose a set of demonstrations to attach notes to, so they do not apply to test-time methods that revise a single output.

\begin{table}[h!]
\centering
\small
\begin{tabular}{lllll}
\toprule
\textbf{Method} & \textbf{Error source} & \textbf{Grounding} & \textbf{Selection} & \textbf{Position} \\
\midrule
LEAP & zero-shot & gold & all & global \\
Self-Refine & test-time output & label-free & -- & -- \\
\method{} & leave-one-out & gold & retrieved & inline \\
\bottomrule
\end{tabular}
\caption{\label{tab:design_space} Existing error feedback methods situated across the four design dimensions.}
\end{table}

\section{Representative Error Feedback}
\label{sec:app_feedback_examples}

Table~\ref{tab:feedback_examples} shows one representative feedback item per dataset, taken verbatim from the manually labeled useful set described in \S\ref{sec:results}. The items illustrate the range of guidance the model produces across task types rather than being sampled by category, so their distribution should not be read as representative of the proportions in Table~\ref{tab:feedback_quality}.

\begin{table}[h!]
\centering
\small
\begin{tabular}{llp{0.62\linewidth}}
\toprule
\textbf{Dataset} & \textbf{Task} & \textbf{Error feedback} \\
\midrule
NCBI & IE & Only include the core disease term in the entity span, and exclude preceding descriptive adjectives unless they are part of the canonical disease name. \\
\addlinespace
TweetNER7 & IE & Government bodies and administrations should be classified as ``corporation'', not ``group''. \\
\addlinespace
PHEE & IE & Only extract \texttt{subject\_disorder} when the text explicitly states that the patient has the condition, not when it is described as an outcome or a risk. \\
\addlinespace
GeoQuery & SP & Use state abbreviations (e.g., ``wa'') in \texttt{cityid} predicates, not full state names. \\
\addlinespace
ATIS & SP & Add the predicates needed to represent the target entity type (``city'') and the relationship (``to'') that links flights to cities. \\
\addlinespace
MTOP & SP & When the sentence instructs the system to send a message or ask someone something, use \texttt{SEND\_MESSAGE} as the intent rather than \texttt{ASK}. \\
\bottomrule
\end{tabular}
\caption{\label{tab:feedback_examples} Representative error feedback generated by \texttt{GPT-4.1} on \texttt{GPT-4o-mini} predictions at a 64-example budget, one item per dataset. IE denotes information extraction and SP semantic parsing.}
\end{table}

\end{document}